\documentclass{article}[12pt]
\usepackage{xurl}
\usepackage{tabularx}
\usepackage{multirow}
\usepackage{subcaption}
\usepackage{tabularx}
\usepackage{wrapfig}

\usepackage{ragged2e}
\usepackage{booktabs}

\newcolumntype{L}[1]{>{\hsize=#1\hsize\RaggedRight} X}

\usepackage[english]{babel}

\usepackage[letterpaper,top=2cm,bottom=1in,left=1in,right=1in,marginparwidth=1.75cm]{geometry}

\usepackage{amsmath}
\usepackage{amssymb }
\usepackage{graphicx}
\usepackage{bbm}
\usepackage[colorlinks=true, allcolors=blue]{hyperref}
\usepackage[nocompress]{cite}
\usepackage{wrapfig,boxedminipage,lipsum}
\usepackage[noabbrev]{cleveref}

\title{TEL'M: Test and Evaluation of Language Models\thanks{This work was partially supported by DARPA under HR001119C0075, AFRL's Autonomous Capability Team 3 (ACT3) and Juniper Networks. Approved for public release; distribution is unlimited.}}
\author{George Cybenko\\ {\small Thayer School of Engineering}\\
{\small Dartmouth College}\\
{\small Hanover NH 03755 USA}\and Joshua Ackerman\\\ {\small Department of Computer Science}\\
{\small Dartmouth College}\\
{\small Hanover NH 03755 USA} \and Paul Lintilhac\\ {\small Thayer School of Engineering}\\
{\small Dartmouth College}\\
{\small Hanover NH 03755 USA}}

\begin{document}
\maketitle
\section{Overview}  
Language Models have demonstrated remarkable capabilities on some tasks while failing dramatically on others \cite{burtsev2023working,wolf2023fundamental,king2023large}.  The situation has generated considerable interest in understanding and comparing the capabilities of various Language Models (LMs) but those efforts have been largely ad hoc with results that are often little more than anecdotal. This is in stark contrast with testing and evaluation processes used in healthcare \cite{FDA1,FDA2,FDA3}, radar signal processing \cite{ciuonzo2015systematic} and other defense areas \cite{DOD1}.
The need for more rigorous test and evaluation of artificial intelligence technologies has been identified as a major challenge \cite{freeman} which arises more broadly in computer science research \cite{cockburn2020threats}.

In this paper, we describe {\em Test and Evaluation of Language Models} (TEL'M) as a principled approach for assessing the value of current and future LMs focused on high-value commercial, government and national security applications. We believe that this methodology could be applied to other Artificial Intelligence (AI) technologies as part of the larger goal of ``industrializing'' AI.

Most existing work aimed at evaluating LMs has been specific to certain classes of tasks and prompts with empirical performance results that are typically based on artificially defined benchmarks and ungrounded in rigorous science \cite{fu2023chain,leaderboard,yu2023skill,zheng2023gpt,wang2023pandalm}.  Little attention is given to rigorous ``Experimental Design.'' Many such results are based on small samples of tasks and experiments with only qualitative summaries of performance.  Moreover, testing on benchmarks that are not aligned with or derived from end-use applications of an LM may not be meaningful to the end-user.  

  \setlength{\fboxsep}{0.5em}

  \begin{wrapfigure}{l}{0.5\textwidth}
        \begin{boxedminipage}{0.5\textwidth} 
        \begin{center} {\bf The Five Ingredients of TEL'M :} \end{center}
            \begin{enumerate}
    \item 
    Identification of {\bf LM Tasks} of Interest
    \item 
    Identification of Task {\bf Properties} of Interest
    \item 
    Identification of Task Property {\bf Metrics}
    \item 
    Design of {\bf Measurement Experiments}
    \item 
    {\bf Execution  and  Analysis} of Experiments
\end{enumerate}
        \end{boxedminipage}
\caption{Many existing language model evaluations fail to follow or document these simple steps.}  \label{wf:1}
    \end{wrapfigure} 
    
TEL'M consists of five ingredients (see Figure \ref{wf:1}) that will be described in detail below.
While these ingredients appear straightforward, they can be nuanced and often nontrivial in the context of evaluating LMs for certain tasks and properties, especially with respect to measurement and analysis.  

A simple example is provided in the Appendix to illustrate and ground the concepts. The example is meant to be ``soup to nuts'' in the sense that motivated readers can 
recreate the datasets, language model architecture, training and experiments with minimal effort.  More complex examples are being documented for future publication.

The main contribution of this project is a novel taxonomy of language model properties and a rigorous methodology for measurement and analysis of those LM task properties.  

After a brief review of background information, we describe the TEL'M methodology and its ingredients in detail with the running example.  Because language models and AI in general are currently such active research areas, we cannot keep up with all the recent contributions and apologize if we have omitted relevant work in this paper.

\section{Background} 

It is assumed that readers are already familiar with Language Models of various flavors such as:
\begin{itemize}
    \item 
    Transformer-based Language Models (currently the most promising and studied LMs) \cite{zhao2023survey};
    \item Multimodal Foundation Models such as Blip-2 \cite{li2023blip} and CLIP \cite{radford2021learning};
    \item 
    Auto-regressive Language Models \cite{brown2020language,malach2023auto};
    \item 
    Recurrent Neural Network Language Models \cite{xiao2020research}; 
    \item State space language models \cite{gu2023mamba};
    \item 
    Hybrid Models \cite{dai2019transformer}
\end{itemize}
as well as the current and proposed use cases and the various technologies underlying them \cite{sherstinsky2020fundamentals,vaswani2017attention,ahmed2023transformers}. 

There is growing interest in LM performance and benchmarks \cite{bommasani2023holistic,jha2023dehallucinating,chang2023survey,wei2022emergent,settles2009active,zou2023representation,burtsev2023working,wolf2023fundamental,king2023large} with recent acknowledgement that this is a hard problem \cite{arvind2023}.
Many suggestions are proposed in the commercial literature \cite{ceylan2023} and a large number of benchmark-based methods have surfaced (Big Bench \cite{srivastava2022beyond}, GLUE Benchmark, SuperGLUE Benchmark, OpenAI Moderation API, MMLU, EleutherAI LM Eval, OpenAI Evals
Adversarial NLI, LIT, ParlAI, CoQA, LAMBADA, HellaSwag, LogiQA, MultiNLI, SQUAD to name a few).  A review of existing approaches demonstrates that they are not quantitative or rigorous enough to past muster with respect to accepted testing requirements \cite{dod5000.89,ammann2016introduction}. In particular, existing use of benchmarks do not investigate the extent to which a benchmark can predict or quantify certain properties on future prompts (that is, statistical soundness of any conclusions) and do not identify factors affecting performance dependence as would be possible with more rigorous experimental design and test execution.

LMs can be black box, gray box or white box according to the visibility into the architecture and training data used to create an 
LM (see Table \ref{table1}). Remote Black Box LMs typically throttle the number of prompts so sustained access for testing could be difficult unless priority access to an API is given.  For example, ChatGPT limits users to a small number of free prompts but allows unlimited prompts on its subscription option. Additionally, reproducability may not be guaranteed because of randomness in the response generation and/or continuous adaptation of the LM platform.
    \noindent
    \begin{center}
    \begin{table}
\begin{tabular}{|c||p{3.15cm}|p{3.3cm}|p{6cm}|} 
    \hline 
     {\bf LM Type}    & {\bf Description} & {\bf Examples} & {\bf Different Use Cases}\\ \hline \hline
    White Box     & Open source code; \newline User training data;\newline User trained.  & Sagemaker \cite{AWS}; \newline nanoGPT \cite{nanogpt}.
    &  Can we use internal representations \newline of facts and concepts to measure computational properties? \\ \hline
       Gray Box  & Foundation models; \newline
       Visibility into  architecture but
       not into pre-training. & Llama \cite{Llama}; \newline
       BERT \cite{bert}; \newline  OpenAI Davinci \cite{davinci};\newline BLIP-2 \cite{li2023blip} \newline CLIP \cite{radford2021learning}
               & How does pre-training affect mission performance?
               \newline Did fine-tuning overcome biases in the pre-trained model?\\ \hline
      Black Box   & Fully remote LM; \newline
      Competitor LM. & Baidu ERNIE \cite{BERNIE}; \newline Yandex \cite{yandex}; \newline ChatGPT \cite{chatgpt}.      & How does LM $A$ compare with LM $B$? \newline What are architectural features of an LM?\\ \hline
\end{tabular}
\caption{ There are several categories of LMs with varying possible use cases and observability. In White Box LMs, questions might be focused on various aspects of performance, trust and alignment. On the other hand, Black Box models could be of interest for comparisons and reverse engineering of  LMs' structures.} \label{table1}
\end{table}
\end{center}


\subsection{Common Inadequacies Made by Many LM Testing and Evaluation Efforts}

We have come across several types of problems arising in published work purporting to test and evaluate language models.  The inadequacies we have found are enumerated in Table \ref{inadequacies}.
We do not indict specific authors or works by attributing possible inadequacies they may have made according to our reading of the publications.  We encourage readers to make their own judgements when reviewing publications that purport to evaluate language models.

    \noindent
\begin{center}
\begin{table} 
\begin{tabularx}{\textwidth}{|L{0.8}|| L{1.9} | *{3}{L{1.3}|} }

    \hline 
     {\bf Inadequacy}    & {\bf Description} &   {\bf Possible Remedies}\\ \hline \hline
    Training-Testing Semantic Mismatch   & An LM trained to perform Task A but is tested for performance on semantically difference Task B  & Align semantics of training and testing with tasks of interest
    \\ \hline
    Training-Testing Distribution Mismatch   & An LM trained to perform a task is tested for performance on semantically similar test data but the test data is drawn from a different distribution than the training data& Check and align test data with either expected end-use data distribution or training data distribution
    \\ \hline
    Tested samples drawn from open source benchmarks & Public benchmarks could be used for training a foundation model and/or fine-tuning a proprietary model so results may not represent intended use-case performance&  Use custom test data aligned with end use case and/or understand if any benchmarks were used in training and/or tuning
              \\ \hline
       Tested on too few samples & Not enough test cases used to obtain statistically significant results &  Use more samples based on statistical analyses
              \\ \hline
      No confidence or error bounds for results   & 
      Reporting results with no expression of statistical confidence or error bounds & Apply well-known and documented statistical procedures to get bounds \\ \hline
      Uninterpretable metrics &
      Results do not align with meaningful or interpretable assessments of interest to an end user  & 
      Align properties and metrics with intended domain  \\ \hline
       Quality of ``ground truth'' is suspect & The outputs of an LM is are compared with ``ground truth'' whose quality is not established or reported & Estimate and report the quality of ground truth used in testing 
    \\ \hline
        Experimental details not included &
      Details of datasets, testing, evaluation or other data not reported  & 
      Specify details of experiments sufficient for reproducability  \\ \hline
\end{tabularx}
\caption{There are multiple ways in which the testing and evaluation of language models can be faulted.  This table shows several major inadequacies with suggestions for possible remedies.}
\label{inadequacies}
\end{table}
\end{center}


\section{The TEL'M Methodology}
Steps in the TEL'M methodology as listed in Figure \ref{wf:1} will be described in detail below and with concrete examples based on a simple paritycase-study example in the Appendix.  To refresh and prepare the reader we list the steps again.
\begin{enumerate}
    \item 
    Identification of {\bf LM Tasks} of Interest:  The problem(s) that  the LM is expected to solve.
    \item 
    Identification of Task {\bf Properties} of Interest:  The properties that tasks (that is, problem solutions) are relevant to test and quantify.
    \item 
    Identification of Property {\bf Metrics}:  The manner in which the properties' metrics are quantified and measured.
    \item 
    Design of {\bf Measurement Experiments}: The design of experiments that are to be conducted to estimate the metrics and the statistical methodology used to analyze results.
    \item 
    {\bf Execution  and  Analysis} of Experiments: The actual execution of the experiments and subsequent analyses of results.
\end{enumerate}
We recognize that following this methodology sequentially will be aspirational in many LM test and evaluation situations because, in spite of best intentions, unforeseen experimental and analysis issues can arise. Nonetheless, people who conduct LM test and evaluation as well as consumers of such results should be aware of experimental science pitfalls such as``$p$-hacking'' \cite{head2015extent}.

\subsection{LM Tasks}

LMs have demonstrated abilities on a wide range of tasks, from concrete and relatively objective, such as authoring computer programs \cite{fan2023large}, to highly subjective, such as authoring creative prose or poetry \cite{franceschelli2023creativity}.  While it is technically interesting to ask if an LM trained on a generic corpus of English language can effectively respond to arithmetic \cite{yuan2023well} or planning \cite{singh2023progprompt}
prompts, in general we assume that most end-user concerns  involve LMs that are trained to perform specific kinds of domain-specific tasks and that rigorous test and evaluation of those LMs will be done on samples of the same kinds of tasks.

Accordingly, it is standard practice to consider LM-based tasks that are consistent with the design and training of an LM, with the general exception of deliberately testing out-of-distribution (OOD) performance on tasks for which the LM was not specifically trained. With this understanding, example tasks that would be appropriate for test and evaluation are, for instance:

\begin{itemize}
\item 
    {\bf  Example: TASK A}: Use text and image prompts in a multimodal language model to answer questions about the image, to caption images semantically and/or synthesize novel properties of the image;
    \item 
    {\bf Example: TASK B}: Synthesize programs that implement APIs for a complex cloud computation whose components are user or third party sourced;
    \item
    {\bf  Example: TASK C}: Integrate symbolic reasoning with neural processing to answer natural language prompts that produce quantitative and logically consistent military Courses of Action that are presented in a predefined, structured Document Definition Language (DDL).
\end{itemize}
\vspace{0.01in}

\noindent
Ideally, Tasks of Interest should be defined well enough to anticipate the properties one might ask of an LM performing that task.  A Task of Interest should have at least these characteristics:

\begin{itemize}
\item 
    {\bf Concreteness}: The Task of Interest should be describable at a high semantic level in prose for formulating hypotheses about the LM's performance. Examples of desired performance should be documented and explained clearly. For example, the tasks of ``knowing a lot of facts'' and ``doing my job'' are not concrete enough for meaningful testing;
    \item 
    {\bf Consistency}: The task should be consistent with the design, training data and actual training of the LM, although as noted above, it can be reasonable to explore those boundaries.  For example, the task ``Play Miles Davis's Kind of Blue'' cannot be done by ChatGPT because of copyright issues and the LM's limited output modality;
    \item
    {\bf Observability}: The task should be one for which an end user will be able to quantitatively or qualitatively assess at least some meaningful properties of LM responses that can be used for assessing task performance. In case the task is related to inferring some aspect of an LM's architecture or training data, a property, for example the number of attention layers, should be observable in the sense of classical control theory \cite{liu2013observability};
    Examples of such properties are whether the model is auto-regressive or uses causal masking. The observability of a property may be LM specific because vendors of commercial LMs may not want certain details of the implementation observable and might add barriers or obfuscation techniques.
\end{itemize}
Part of the current excitement about LMs is the exploration of tasks which they are either capable or incapable of doing so this will likely be a dynamic area for some time, as the technology and expectations evolve.

\subsection{Task Properties}
Language Models operate by accepting an input prompt, which can be multimodal, and
producing a response, which can also be multimodal.  More formally, and to establish our notation, we regard LMs to be functions mapping an input prompt space, $\cal{X}$, to an output response space, $\cal{Y}$. We will denote a specific, fixed LM by $\cal{L}$ so that $\cal{L}:X \rightarrow Y$ and ${\cal L}(x) =y$ is the response to prompt $x$.  

In many cases, the output of an LM depends on various probabilistic ingredients (such as sampling disciplines that depend on random number generators that typically are not controllable by end users)  so that ${\cal L}(x)$ is possibly random and not always reproducible.  That is, repeatedly prompting an LM with the same prompt may not always produce the same response as output. We will revisit this aspect of LMs once we start discussing property metrics below.

By their very nature, different LM tasks will have have different properties suitable for test and evaluation. Example properties of LM tasks are:

\begin{itemize}
 \item
  {\bf Accuracy}: Accuracy of an LM, and any AI in general, is often a desirable property. In classical information retrieval, there are several ways to define accuracy \cite{clough2013evaluating}.  On the other hand, if a task for 
  an LM involves creativity, alignment with facts may not be important.
  This type of property is discussed in more detail in the metrics subsection.
   \item
  {\bf Sensitivity}: Task sensitivity captures the extent to which
  perturbations in the inputs lead to changes in the outputs. This could be sensitivity to bit flips in Boolean analysis, or a definition of sensitivity more appropriate for natural language tasks. Task sensitivity has been found to be a good predictor of task performance in both formal languages and natural languages \cite{hahn2021sensitivity}.
  \item {\bf Monotonicity:}
  This property captures whether the performance of the LM is monotonic in the complexity of the task quantified in some objective way \cite{goldreich1998monotone}.  For example, does 
  the LM perform better on shorter prompts?
  \item
  {\bf Training Efficiency}: Significant progress is being made on the training of foundational and fine-tuned LMs \cite{efficiency}. This property has to do with the effectiveness and efficiency of such training with respect to training and fine-tuning.
  \item 
  {\bf Prompt Efficiency}:  So-called ``prompt engineering'' sheds light on the extent to which an LM's ability to effectively perform a task
  can depend on the structure of a prompt.  Chain-of-thought (COT) prompting is an example of this \cite{wei2022chain}. This task property has to do with the extent to which the LM end users have to craft specific types of prompts with specific structures to achieve effective responses.
  \item
    {\bf Explainability}: This is the extent to which responses of an LM on a task can be interpreted and understood by  human users. This property has already been explored in detail for some AI technologies \cite{hoffman2018metrics}.
    \item 
    {\bf Extensibility}: There is growing interest in how well fine-tuning will fare when extending a foundation model to a specific use case and domain \cite{Bommasani}.  It has to do with the size and scope of the fine-tuning used and the subsequent performance improvements.
    \item {\bf Creativity:} ``Hallucination'' is often criticized as an undesirable LM artifact but deviation from factual accuracy and prior training is a distinguishing property of creativity \cite{franceschelli2023creativity}.  In some tasks, a user might assign value to the novelty of an LM response.  In the national security domain, this could arise when novel course of action plans are being sought for a specific tactical situation.  Recent work has explored internal properties of LMs for distinguishing between fact and fiction 
    \cite{azaria2023internal}.
    \item
    {\bf Ethics and Biases}: Ethics and biases are of great concern within the US government \cite{whaiethics} and can be factors across virtually all LM applications. While the concerns are well-placed and appropriate, there are few principled proposals for how to quantify and measure these properties.  TEL'M can investigate these issues and offer approaches for developing objective metrics for such properties.
    \item {\bf Usefulness}:  For some task products, such as LM authored code which may not be accurate or correct, the performance of the LM might still be considered ``useful'' in the sense that the code artifact is the basis for an effective program construct with additional human effort.  For example, an LM produced code could have a high level structure that is initially not correct or executable but provides a template that a human programmer can easily fix \cite{raychev2014code}.
\end{itemize}

We will denote by ${\cal F}$ the set of {\em all} functions mapping ${\cal X}$ to ${\cal Y}$ of which a trained LM ${\cal L} \in {\cal F}$ is but one example.  For properties, $P$,  such as listed above, the subset of functions having the property will be denoted by ${\cal F}_P$ so that ${\cal L} \in {\cal F}_P$ means that ${\cal L}$ has property $P$.  

This is a very general framework for discussing properties. For example, consider the ``correctness'' property.  Assuming there is an objective way to decide whether an individual LM response is correct or not, it may be too strict a requirement that the LM is ``always correct.''  Instead, in many use cases, it is acceptable to be correct ``most of the time'' in which case the relaxed correctness property $P$ is that the LM produces a correct response, say at least 95\%, of the time a random prompt is input.
This itself is a property and ${\cal F}_P$ becomes the set of functions that are correct 95\% or more of the time.  This is discussed in more detail below.

\begin{figure}[h] 
\begin{center}
\includegraphics[width=6.5in]{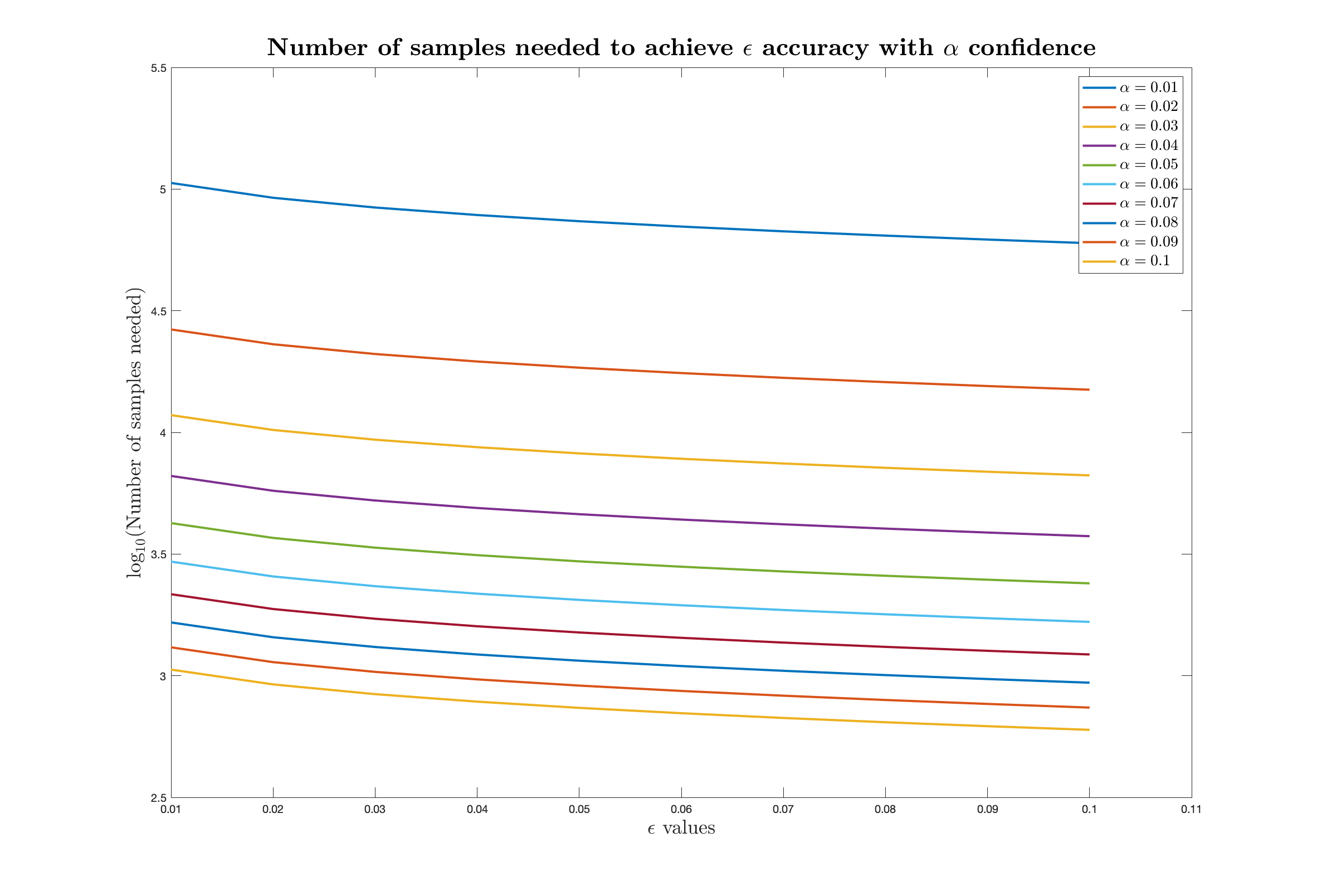}
\caption{Hoeffding's Inequality \cite{hoeffding1994probability} can be used to bound the number of samples needed to get confidence $1-\alpha$ for the probability that a property, $P = P[\mu-\epsilon,\mu+\epsilon]$ as described in the Confidence Intervals (CI) box. For example, to get an estimate of that probability to within $\epsilon=0.1$ with confidence $1-\alpha = 0.95$, we need at least 738 samples.  Constraints and more sophisticated models for the property could significantly reduce that number.} \label{hoeffding}
  \end{center}
\end{figure}

\subsection{Property Metrics}
The role of property metrics is to provide estimates of 
how ``close" an LM, ${\cal L}$, is to having a property, $P$.  To make this more rigorous, we explicitly define several ingredients fundamental to TEL'M notions of properties and metrics, along with the associated notation used here. 
\begin{itemize}
\item 
$D$ is a probability distribution over inputs as determined by the particular use-case and task envisioned for the LM.  This distribution could be formal and explicit (such as the uniform distribution over binary strings of length 100) or it could be entirely empirical (such as the set of prompts presented to ChatGPT in the year 2023).
\item 
$TD_N$ is a ``test data set'' of size $N$, namely 
$$TD_N=\{( x_i,y_i )~ | ~ x_i \in {\cal X}, {\cal F}(x_i)=y_i \in {\cal Y}, 1 \leq i \leq N ~\}.$$  
In general, it is assumed that the $x_i$ are sampled independently from $D$, specifically that $~x_i \sim D$.
However, in cases where the LM responses are not deterministic, it may be desirable and necessary to repeatedly apply ${\cal L}$ to the same random input $x_i$ to get a sample of the possible outputs for the same input prompt.  In other test situations, samples can be selected based on previous samples and their responses which are then now independent samples.
Such sampling is sometimes useful but beyond the scope of this short overview.
\item 
${\cal A}$ is an algorithm or procedure  that is applied to $TD_N$ to provide an estimate of the extent to which the LM has the property $P$.  
\item 
${\cal C}$ is statement about the ``confidence'' in the estimate that ${\cal A}$ will produce a similar estimate when presented with other test data sets, $TD'_N$.
Such confidences are typically expressed in terms of the distribution $D$ imposed on the test data.  That is, a confidence of 0.95 means that 95\% of the time a similar test data set of size $N$ is obtained, a similar estimate will be derived by the algorithm ${\cal A}$.
\end{itemize}

To see how these ingredients fit together, consider an LM, ${\cal L}$, and a task for which ``accuracy'' is the property of interest, $P$.  This assumes we have knowledge of the correct response for input prompts, $x$, so that $T(x)$ are the correct responses for input $x$.  We select $N$ (the value of which will be discussed shortly) and obtain $N$ samples from $D$, $\{(x_i)\}$ where $1 \leq i \leq N$, and evaluate ${\cal L}(x_i)$ to obtain the test data results, $TD_N=\{(x_i,{\cal L}(x_i)\}$.  Comparing ${\cal L}(x_i)$ with the correct responses $T(x)$ defines a Boolean variable, $B(x)=1$ if ${\cal L}(x)= T(x)$ and $B(x)=0$ otherwise.

The expected value or mean of $B(x)$ over $D$ with respect to the underlying probability distribution of input prompts in $D$, is
\begin{equation}\mu = \mathbb{E}_{x \sim D}(B(x)) = \Pr\{x \in D | B(x)=1\}= \lim_{N \rightarrow \infty} \frac{\sum B(x_i)}{N} \label{lln}
\end{equation}
where the mean is the limit shown on the right by the Law of Large Numbers (LLN) (note that the mean and variance of the random variable $B$ are both finite).  This means that the probability of the model ${\cal L}$ being correct for a randomly chosen input prompt, $x$, is $\mu$ and can be approximated by simply computing the fraction of tests that are correct.
This is what the vast majority of LM correctness testing efforts do now.

The rate at which the limit on the right in (\ref{lln}) approaches $\mu$ as $N \rightarrow \infty$ can be estimated using one of several of probability inequalities such as Chebyshev's, Chernoff's or Hoeffding's \cite{hoeffding1994probability}.  In particular, Hoeffding's inequality states that for $0 \leq B(x) \leq 1$ (which is the case here for Boolean $B$)
\begin{equation}
    \Pr_{x_i \sim D}\left(\left| \frac{\sum B(x_i)}{N} - \mu \right| \leq \epsilon \right) \geq 1 - 2e^{-2N\epsilon^2} = 1-\alpha \label{hoeffdinginequality}
\end{equation}
which is a rigorous relationship between $\epsilon$, $\alpha$ and $N$.
Namely, selecting any two of these variable implies a bound on the third.

In terms of the ingredients listed above, we have
\begin{itemize}
\item 
The test input data is drawn independently from $D$.
\item 
$TD_N$ is a ``test data set" of size $N$ that is used to infer aspects (the probability of being correct in particular).
\item 
The algorithm, ${\cal A}$, is the comparison of the LM responses with ``truth" and computation of the ratio between the number correct responses to the total number of inputs tested.  The $\mu$ and $\epsilon$ in inequality (\ref{hoeffding}) is the extent to which ${\cal L}$ is correct.
In particular,  the true but unknown probability that ${\cal L}(x)$ is correct on a random input prompt, $x \sim D$, is the mean value of $B(x)$ over $D$, namely $\mu$, which is  approximated to within a tolerance $\epsilon$ by the test sample average, $\sum B(x_i)/N$.
\item 
The statement about the ``confidence'' in the estimate, ${\cal C}$, is the assertion that a similar estimate when presented with other test data sets, $TD'_N$, will occur with probability at least $1-\alpha$.  
\end{itemize}

\noindent
Solving for $N$ in (\ref{hoeffdinginequality}), we must have
$$N \geq \frac{\ln(2/\alpha)}{2\epsilon^2}$$
for the inequality to hold.  A graphical depiction of the relationship between $\alpha$, $\epsilon$ and $N$ is shown in Figure \ref{hoeffding}.

Once an appropriate number of test samples have been generated, there are several classes of algorithms that could be applied to the test data for property evaluation.  We outline three main techniques below.

\vspace{0.2in}

\begin{figure}[t] 
    \begin{center}
\includegraphics[width=5.5in]{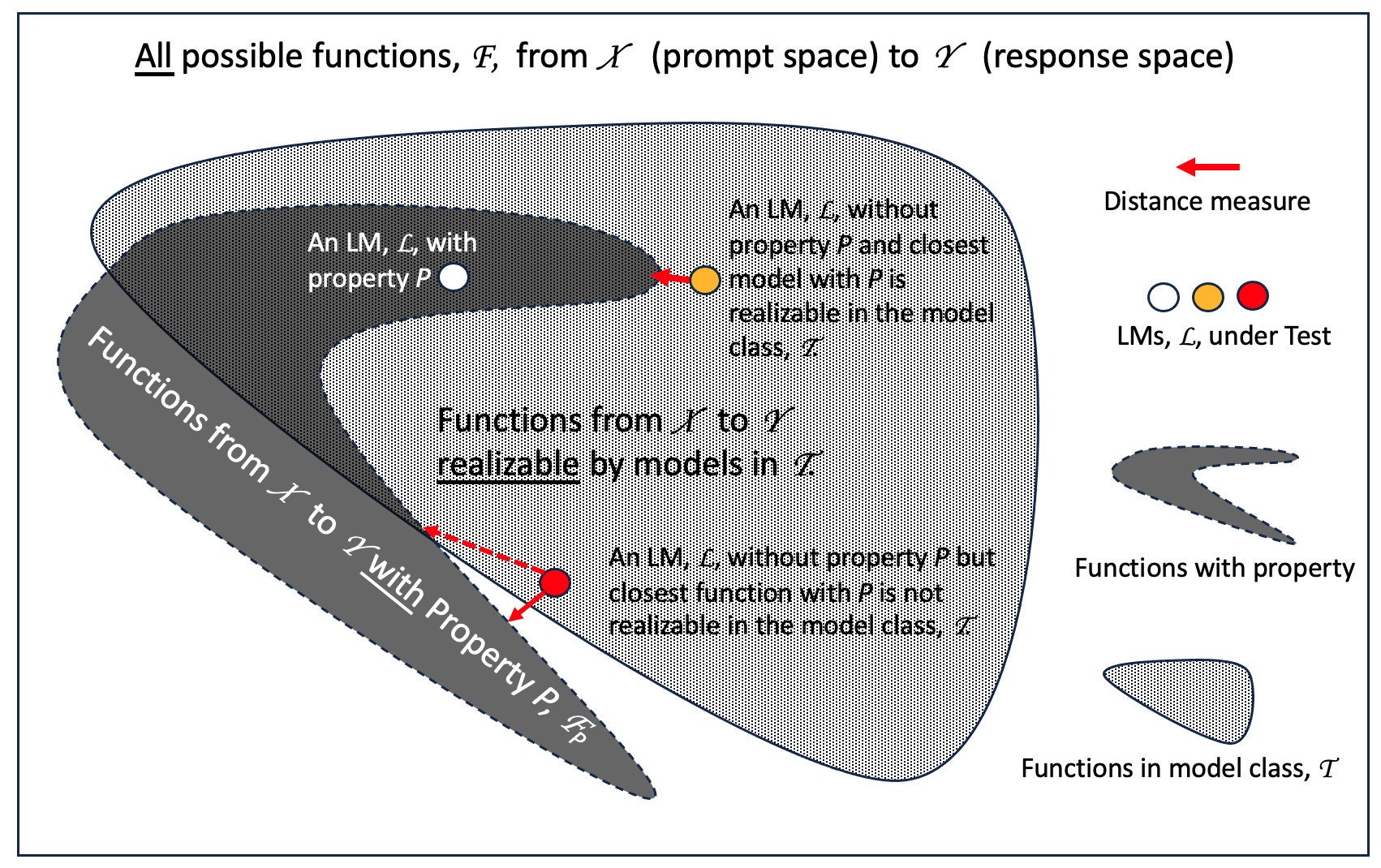}
\caption{Ideally, a property metric would provide some insight into how far the LM under test is from  having the property.  For example, saying that a system is 90\% accurate should translate into meaning that 10\% of the responses would have to be changed to make the system 100\% accurate.  As discussed in the text, this can be translated rigorously into meaning that the system is a distance of 0.1 away from the property. However, the notion of ``distance'' from an LM, ${\cal L}$ to the property under consideration has several nuanced considerations as depicted in this figure. For example, the ``closest" system with the property may not be in the architecture class of language models (such as transformers with the same specific sizes and configurations of attention and feedforward layers as the LM under test, not just the generic class of all transformers) to which the LM under test belongs.  The problem of determining whether any model in that same architectures class has the property is typically a difficult theory oriented question about the representational powers and trainability of language model architectures. } \label{MUTDX1}
  \end{center}
\end{figure}

\noindent
\begin{center}
\fbox{\begin{minipage}{0.9\textwidth}
\noindent
{\bf Confidence Intervals (CI)}:  When tests are used to estimate scalar parameters of an LM, such as accuracy or other properties that can be formulated as averages, some form of confidence interval estimation can be used.  For example, accuracy can be quantified as the ratio of correct responses to all responses.  In general, if a property can be expressed as the average of measurements of the test data (what we call a ``simple property'' as discussed below) and no assumptions are made about that average value, then confidence interval techniques can be applied. Using Law of Large Number type results such as Hoeffding's Inequality \cite{hoeffding1994probability}, it is possible to obtain statements such as 
\begin{equation} 
\Pr({\cal L} \in {\cal F}_{P[\mu-\epsilon,\mu+\epsilon] }) \geq 1-\alpha \label{eq:2}
\end{equation}
where $P[\mu-\epsilon,\mu+\epsilon]$ is the property that that the LM has accuracy in the interval $[\mu-\epsilon,\mu+\epsilon] \subset [0,1]$.  Given an $\epsilon$ and $\alpha$, Hoeffding's Inequality can be used to estimate how large $N$, the test data size, should be.  A graphical depiction of the relationships between $\alpha$, $\epsilon$ and $N$ is shown in Figure \ref{hoeffding}.

It is important to note that the method of confidence intervals is relevant when no prior assumptions about the property being tested are made except that the test samples are randomly drawn and when a probability distribution of the test statistic can be estimated.
\end{minipage}}

\vspace{0.2in}

\noindent
\fbox{\begin{minipage}{0.9\textwidth}
\noindent
{\bf Classical Hypothesis Testing (CHT)}: Unlike the method of confidence intervals, hypothesis testing begins with a hypothesis about some parameters of the models being tested.  For example, we may want to know if an LM, say ${\cal L}$ with accuracy $\mu$, is more accurate than another LM or system for performing the same task, say ${\cal L'}$.  If ${\cal L'}$ has known accuracy $\mu'$, then we are asking if the accuracy, $\mu$, of ${\cal L}$ satisfies
$\mu \geq \mu'$
which is a ``hypothesis'' about $\mu$ and this question becomes the hypothesis.

Classical hypothesis testing has been a standard approach for analyzing hypotheses about systems based on experimental data for over a century  \cite{ciuonzo2015systematic}.  We will not review those concepts here and instead point readers to review articles focused on the application of CHT to machine learning which would be most relevant to this paper \cite{emmert2019understanding}.

It is important to note that CHT is relevant when some assumptions about the property being tested are made and when a probability distribution of the test statistic can be estimated or postulated. In spite of CHT's long and established history, there are ongoing critiques of CHT \cite{greenland2016statistical}.
\end{minipage}}

\vspace{0.1in}
\noindent
\fbox{\begin{minipage}{0.9\textwidth}
\noindent
 {\bf Property Testing (PT)}:  Another paradigm has emerged in the past few years for evaluating hypotheses about properties of functions.  Called ``Property Testing", this approach was developed to handle situations for which confidence interval, classical hypothesis testing and other classical statistical methods that require knowledge of a test statistic's probability distribution may not be easily computed and therefore cannot be directly applied \cite{goldreich1998property,goldreich2017introduction}. 

As above, define $D$ is the probability distribution over which samples of the task will be drawn and $P$ is the property being tested.  We say that the model function ${\cal L}$ is $\epsilon-$far from having property $P$ when 
$$\min_{h \in {\cal F}_P }\left\{\Pr_{x \sim D}(h(x) \neq {\cal L}(x)) \right\} \geq \epsilon.$$ 
In other words, ${\cal L}$ is $\epsilon$-far from ${\cal F}_P$ means at least an $\epsilon$ fraction of ${\cal L}$'s responses would need to be changed in order to match the closest $h \in {\cal F}_P$, that is, to have the property. As depicted in Figure \ref{MUTDX1} the closest function with the desired property may or may not be in the language model architecture class under consideration.

Within this framework, an algorithm ${\cal PT}$ is called a ``property tester'' for property $P$ if, given an $\epsilon \geq 0$ and a test data set $TD_{N(\epsilon)}$ of size $N(\epsilon)$ (that is, the test data size depends on $\epsilon$), it satisfies these two conditions:
\begin{itemize}
     \item If ${\cal F} \in {\cal F}_P$ then 
         \[
    \Pr_{TD_{N(\epsilon)}} \left\{  {\cal PT} ~\text{applied to } ~TD_{N(\epsilon)}\text{ will conclude that } {\cal F} \in {\cal F}_P ~~\text{correctly}\right\} \geq \frac{2}{3};
         \]
 \item If ${\cal F}$ is $\epsilon$-far from ${\cal F}_P$ then 
 \[
    \Pr_{TD_{N(\epsilon)}} \left\{  {\cal PT} ~\text{applied to } ~TD_{N(\epsilon)}\text{ will conclude that } {\cal F} \notin {\cal F}_P ~~\text{correctly}\right\} \geq \frac{2}{3};\]
 \end{itemize}
An effective property testing algorithm requires that $TD(\epsilon)$ is reasonable and that the run time complexity of the testing algorithm ${\cal PT}$ is ``efficient."  It is important to note that the property testing paradigm is silent about the situation in which the model, ${\cal L}$, is neither in ${\cal F}_P$ nor $\epsilon$-far from ${\cal F}_P$ which might be a large part of the potential model space.  

We refer readers to recent publications that develop the concepts and recent results in more detail \cite{goldreich2017introduction}.

\end{minipage}}
\end{center}

\subsubsection{Simple Property Metrics}

We call properties ``simple'' if the test data for the property consists of individual independent prompt-response samples which are individually scored and then averaged to obtain the property metric.  Hoeffding's or similar probabilistic inequalities can be used to estimate the required test sample sizes to obtain these property metrics with the required accuracy and confidence, as discussed above in the Confidence Bounds description.  We note that any property of a task that is a limiting average of the some point-wise property defined for each input is thus a ``simple property'' by definition.  

All simple property metrics can be derived using confidence interval (CI) estimation or classical hypothesis testing (CHT).
Note that the failure rate of the test can be regarded as the distance to some function having the property. However, in the general context of language models, it is not known in general if the closest function is in the same architectural class (for example a transformer
with specific hyperparameters) as the LM under test or outside that class.

\begin{figure}[t] 
    \begin{center}
\includegraphics[width=5in]{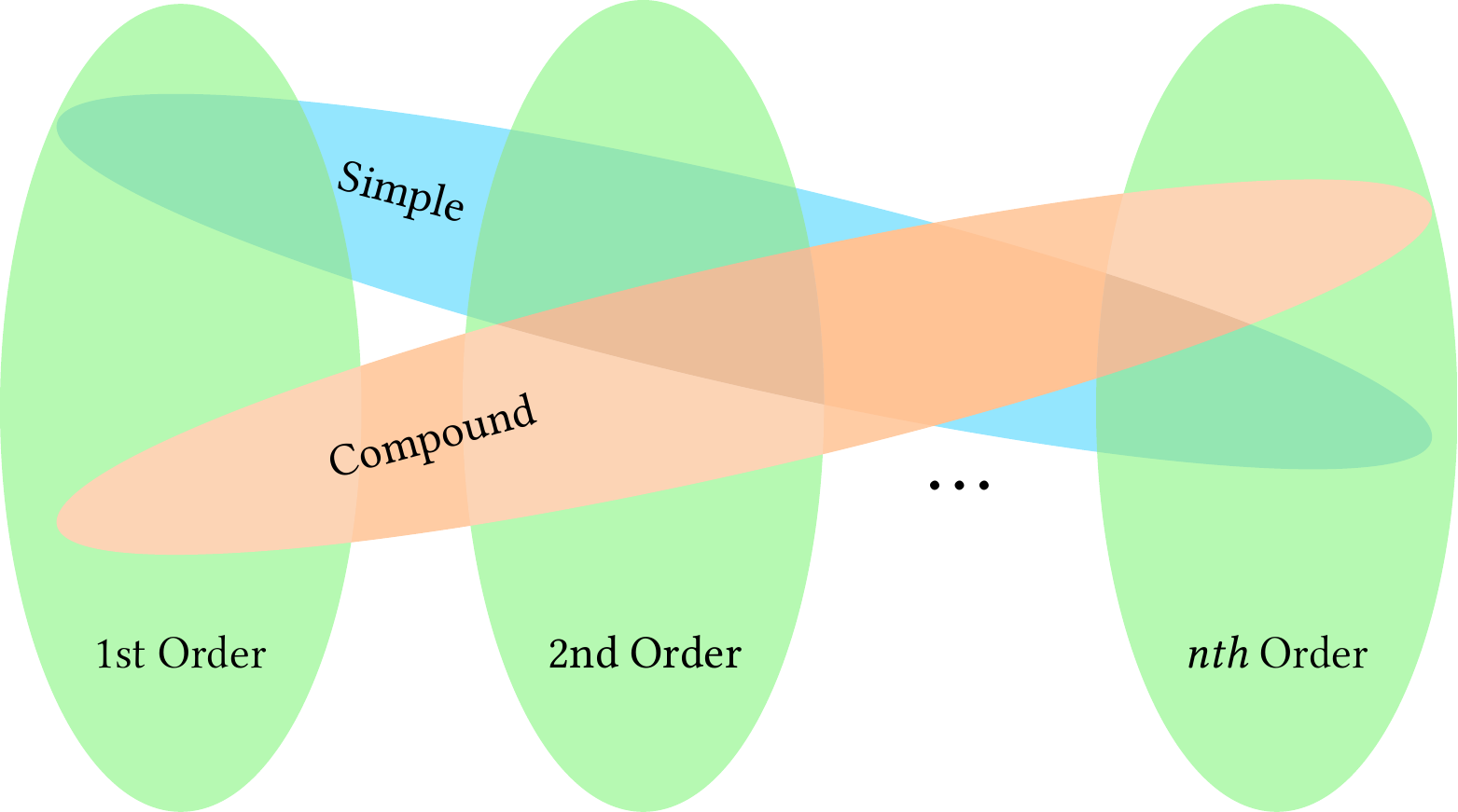}
\caption{Simple task property metrics can be determined by independent samples of an LM's prompt-response behavior that are aggregated by averaging.  By contrast, compound task property metrics, such as monotonicity, require a multiplicity of prompt-response pairs that are not simply averaged although subsequent processing of averages may be needed to make inferences about a compound property. Another dimension along which task property metrics vary is their ``order''.  A first order property can be determined \text{prima facie} from the prompt-response pairs by averaging. On the other hand, second and higher order properties defer to other applications (possibly other LMs, parsers, compilers, optimization packages, execution environments, etc. for example) to estimate  metric values. Those other applications themselves have to be tested and evaluated with metrics that can be simple or compound as this figure depicts.  While 1st order property metrics are either simple or compound, higher order property metrics can be combinations of both.}
  \end{center}
\end{figure}

\subsubsection{Compound Property Metrics}
As defined above, simple property metrics are based on individual prompt-response samples which are aggregated as averages allowing for the use of Hoeffding's or similar probabilistic inequalities.  Compound properties on the other hand are meaningful only in the context of multiple crafted prompt-response samples by comparing and aggregating prompt-response samples for certain sets of samples.  

One such property is ``monotonicity'' of accuracy for example.  Suppose we want to test whether an LM's accuracy, which is a simple property, decreases as prompt complexity, quantified in some principled way, increases.  If $\mu(n)$ is the simple property accuracy for prompts of length, $n$, then we are interested in the whether the relation $\mu(n) \geq \mu(m)$ holds
for all $m > n$. While we can estimate whether $\mu(n) \geq \mu(m)$ using CHT, the monotonic property at large requires combining multiple such tests using additional probabilistic reasoning.  In particular, if we determine that $\mu(n) \in [a,b]$ with probability at least $1-\delta_1$ and $\mu(m) \in [c,d]$ with probability at least $1-\delta_2$ then we still need to estimate the probability that $\mu(n) \geq \mu(m)$ given these measurements.  Property testing for such compound property metrics is an ongoing research area \cite{goldreich1998monotone,langsec2024}.

Estimating an LM's distance to monotonicity with respect to a property metric can be done using a simple linear program as follows. Let the sequence of sample averages of a simple property be $\{\widehat{\mu}_1,...,\widehat{\mu}_n\}$ where $\hat{\mu_i}$ is the estimated property value for. 
Introduce slack variables $\{s_1,...,s_n\}$ and $\{r_1,...r_n\}$, where  $r_i,s_i \geq 0$, and find the minimum 
$$\epsilon = \min_{s_i,r_i}\left[\sum_i \alpha_i(s_i+r_i)\right]$$
subject to the linear inequality constraints:
$\widehat{\mu}_i+s_i-r_i \geq\widehat{\mu}_{i+1}+s_{i+1}-r_{i+1}$ that define monotonicity and $-\delta_i \leq s_i-r_i \leq \delta_i$ that forces $\widehat{\mu}_i+s_i-r_i$ to be in the confidence interval for the estimated $\widehat{\mu_i}$ as discussed above. Note that either $r_i=0$ or $s_i=0$ or both must hold at the optimal solution. The $\alpha_i$ are the fixed and known or assumed probabilities or fractions of the input space that have complexities $i$ so that $\alpha_i \geq 0$ and $\sum_i \alpha_i = 1$.  They are not variables in the linear program.

If this linear program has no solution, it shows that there are no values within the multiple confidence intervals that result in a monotonic metric.  On the other hand, if there is a solution, then at the optimal solution, $s_i-r_i$, is  the minimum (weighted) amount that the estimated $\widehat{\mu_i}$  needs to change in order for the property metric to become monotonic. If the underlying property is simple, such as accuracy, then we can interpret $s_i-r_i$ as the fraction of responses for complexity $i$ prompts that have to be changed for monotonicity
to hold and with the weighting $\alpha_i$, $\sum_i \alpha_i(s_i+r_i)$ is the overall fraction of responses across all complexities.

Another compound property of increasing interest is the sensitivity of a language models' responses to small perturbations in prompts \cite{hahn2021sensitivity}.  Testing for sensitivity requires sampling a neighborhood of a base-point prompt and processing the associated responses to gauge the variability of those responses with respect to the base-point.  

\subsubsection{Higher-Order Property Metrics}
In some LM use cases, the evaluation of LM responses to estimate property metrics  requires nontrivial post-processing itself.  For example, if we are interested in LM accuracy, we need a way to estimate ground truth -- that is, what is ``correct.''  

If the task is a basic arithmetic problem, such as answering ``What is the sum of 25 and 48?'', we can easily have high confidence in the computed, objective correct answer and so can score the response easily. On the other hand, if the task involves natural language prompts and responses corresponding to a question answering task \cite{rajpurkar2016squad} for example, then natural language responses have to be evaluated which means that some confidence in the accuracy of the ground truth is needed, as well as any automated comparison with a natural language response and that ground truth.  This has typically been done by using a heuristic score such as BLEU \cite{papineni2002bleu} or by human scoring of natural language responses (which may not be scalable to large, novel test samples).

As another example, consider the property of writing ``correct" computer code based on natural language descriptions of the desired code.    Testing this property requires checking the program against ground truth which begs the question. 
For example, suppose we ask an LM to write a program to solve certain types of  planning problems \cite{jha2023dehallucinating}.
Programs for solving those types of problems might only produce approximate solutions
because some aspect of the problem involves solving NP-Hard optimization problems.  
In this case, we want to have a metric on how close the response solutions 
are to computing optimal solutions on instances of the planning problem. Accordingly, we would want to quantify the LM performance on this task by asking if the LM
produces planning solutions that come to within say 90\% of the optimal solution 95\% of the time it is presented a problem instance, and that the LM succeeds in doing this 80\% of the time.

For example, consider a task  for which a model is compared with human responses, attaining $p$ accuracy meaning that approximately a proportion of $p$ responses are scored as correct and so $1-p$ are scored as incorrect with respect to human responses.  However, human responses are not in general perfect \cite{freitag2021experts,dodge2017study,barlow2004accuracy}. To explore the implications of this, we introduce these concepts:
\begin{itemize}
\item 
The {\bf Human Accuracy}
 is the accuracy, $q$, of the humans performing the task as measured by comparison with some ``truth" which is typically unknown but might be inferred from other sources (although this is sometimes measured \cite{barlow2004accuracy});
    \item The {\bf Test Accuracy} of the model is the accuracy, $p$, of the model measured by comparison with the test data of which only $q$ fraction is actually correct and;
    \item The {\bf True Accuracy}, $r$, of the model is the accuracy of the model measured against the unknown true results.
\end{itemize}

\begin{figure}[ht] 
    \begin{center}
\includegraphics[width=6.5in]{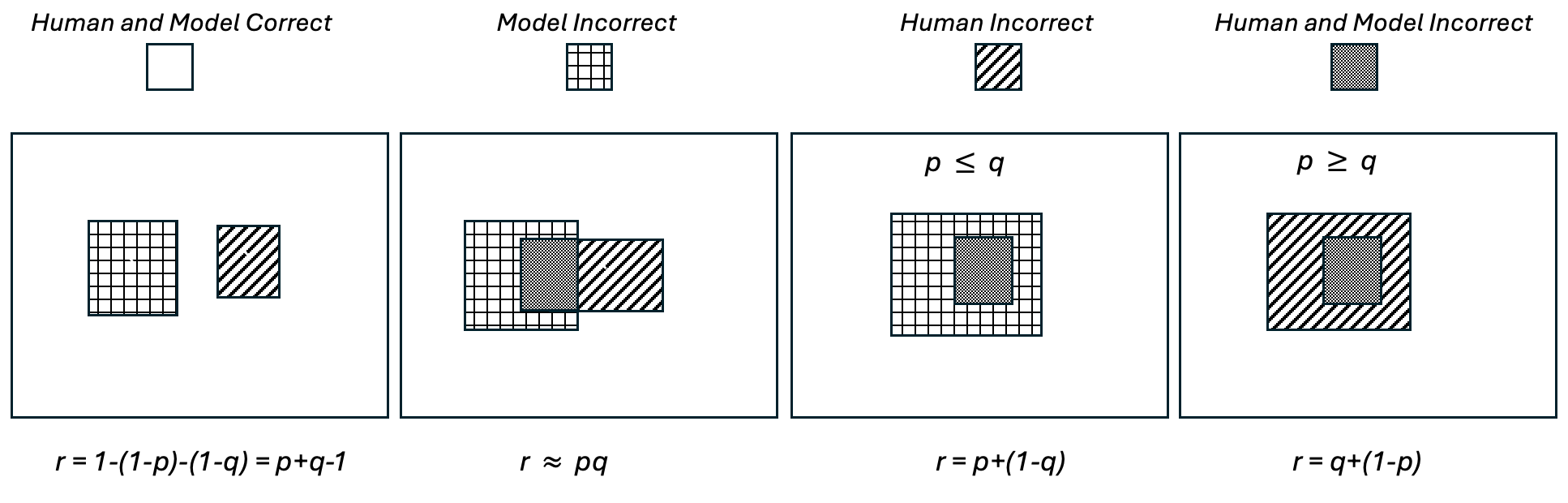}
\caption{This illustrates the concepts and derivations surrounding various possibilities for actual accuracy, $r$, of an LM in a higher-order testing scenario when the data being tested against has true accuracy $q$.
In this context, when both the human and the model, which is scored again the human, are ``wrong'' the model response is actually correct. For example, if $p=0.9$ is the estimated model accuracy measured against data that has accuracy $q=0.95$ then the true value, $r$, will satisfy $0.9+0.95-1 = 0.85 \leq r \leq 0.9 + (1-0.95) = 0.95$. Under the independence assumption for model and human errors, $r=0.9\cdot 0.95=0.855$ which is almost at the lower bound for $r$.} \label{secondorder}
\end{center}
\end{figure}

If the correctness of human and LM responses to individual prompts are independent random variables, then $r=pq$.  However, if they are not independent, then it is possible that most of the model test errors relative to human true errors are aligned.  That is, a model response is scored as incorrect relative to the human response but seeing as the human response is wrong, the model response is actually correct.  So in an extreme case, if $1-p \geq 1-q$ then it is possible that a fraction $1-q$ of model responses that are scored in testing as wrong could in fact be correct because the human generated test responses are wrong on those samples.  In this extreme case, a $p+(1-q)$ fraction of the model's responses would be correct.  

By similar reasoning, if  $1-p \leq 1-q$, then all wrong model responses are on samples for which the human responses are also wrong so we would have $r=(1-p)+q$.

Finally, the other extreme case is when the wrong answers of the model are disjoint from the wrong human answers and we get $r=1-(1-p)-(1-q)=p+q-1$.  Note we assumed that $0.5 \leq p$ and $0.5 \leq q$ in this binary classification problem which can be obtained by flipping the response categories if needed.

Putting these together we get
$$p+q-1 =p-(1-q) \leq r \leq 1-p+q =q+(1-p) ~\text{if}~1-p \leq 1-q$$ and $$p+q-1 =p-(1-q)  \leq r \leq 1+p-q =p+(1-q)~\text{if}~1-p \geq 1-q.$$
(Note that $r=pq$ lies between these upper and lower bounds because 
$$1-p+q-pq=(1-p)(1+q) \geq 0 \geq p+q-pq-1 = -(1-p)(1-q)$$
if $p \geq q$ with a similar derivation for the case $p \leq q$ but with the left term being $(1-q)(1+p)$. ) Figure \ref{secondorder} illustrates these derivations.

Explainability, creativity and usefulness, among other properties, that are very subjective. For test and evaluation at scale, these properties would likely have to be evaluated using tools developed by third parties or human experts whose accuracy would ideally be quantified  accordingly.

 In summary, a higher-order LM property metric  depends on downstream property metrics of responses that themselves have to be tested and evaluated. We believe that the above derived intervals for properties such as accuracy can be extended to $n^{th}$-order metrics as well.

\section{Design of Measurement Experiments}

We have described the concepts of an LM task, task properties and various metrics and algorithms for quantifying the extent to which tasks have those properties.   Although the design of experiments is listed as the second to last ingredient, it should in practice be considered at a high level at the very beginning of a TEL'M implementation.  When possible, some form of  design of experiments methodology should be applied \cite{antony2023design,telford2007brief}. 

Historically, a significant driver for early experimental design work was motivated by the need to systematically control factors that could be used to explain observed results.  Factors in LM experimental design could be the use or not of chain-of-thought prompting, the complexity and length of a prompt within the class of possible prompts for that task, the distribution of prompts over the space of possible prompts, including in-distribution vs out-of-distribution sampling,  among many others.

Ideally, TEL'M ingredients and especially the experiments would be negotiated among the end-users, the LM authors, provider or vendor and the TEL'M test and evaluation team.  

 In summary, a  goal of experimental design is to frame experiments so that the variations in property metrics (the dependent variables)
can be rigorously understood and quantified in terms of variations in features of the prompts (the independent variables) which can include the underlying LM platforms.

  The designs should be documented and reported together with any subsequent deviations from the initial designs.
  For example, we have come across LM evaluations (to remain anonymous) that report  evaluations based on public benchmarks without reporting whether those benchmarks were part of a training or tuning data set.  Subsequent evaluations using test samples not from the benchmarks (so out-of-distribution with respect to the training data) find significantly different performance.  That could be a result of ``$p$-hacking'' \cite{brodeur2020methods} or chance of course, the reality known only to the report authors if not reported.

\section{Execution  and  Analysis of Experiments}
Once experiments have been designed, they have to be executed and the resulting data analyzed in a principled way suitable for documentation, dissemination and if needed, reproduction. A critical component of LM research and evaluation continues to be appropriate access to the computing, storage and data required for training and testing.  Commercial language model developers use resources that are typically difficult for outside, especially academic, evaluators to match or even approximately approach.

Prior to conducting the actual computational experiments, it is appropriate to address and document the following language model characteristics in play during execution:
\begin{itemize}
\item 
Is the LM under test deterministic or stochastic.  That is, do repeated identical prompts produce the same outputs or are they stochastically generated and therefore responses to a single prompt may not be reproducible by the same or other testers.
\item 
Is there a ``temperature" setting?  What is it and can it be controlled by the LM's end-user?
\item 
Some LMs maintain a context of previous prompts.  Does the LM under test have a context, can it be cleared, set or observed during testing?
\item Is the LM static or still being trained? This is especially important in remote Black Box testing of LMs whose training could be continuous and/or based on response feedback mechanisms so that tests may not be reproducible over time.
\item 
Has the LM under test been trained on any benchmarks being used in testing?  In a black or gray box setting, this could be tested by comparing benchmark performance on bespoke, novel  prompt-response samples.
\end{itemize}
 We have provided a straw man form for reporting LM test and evaluations in the Appendix.  
 
\section{Summary}

Rigorous testing and evaluation has been standard practice in industries such as aerospace, pharmaceuticals, food, automotive and  consumer appliances to name a few.  While the Turing Test was proposed early in the AI lifecycle \cite{french2000turing}, recent advances in and applications of AI, specifically language models and related generative systems, significantly expand the scope, scale and complexity needed to evaluate modern AI systems in real-world applications.  We have proposed an initial framework for conducting test and evaluation of language models which we believe can mature to be more relevant and useful as the field matures.

We consider this to be a work-in-progress with the expectation that feedback, constructive criticism and implementations of this framework will constantly improve the scope and quality of the ideas presented here.

\section{Acknowledgments}
This work was partially supported by DARPA under HR001119C0075, AFRL's Autonomous Capability Team 3 (ACT3) and Juniper Networks.  We thank colleagues for discussions, critiques and helpful suggestions.
\bibliographystyle{plain}
\bibliography{LLM.bib}

\newpage

\newpage

\section*{Appendix - Illustration of TEL'M for Parity Problems }

\subsection*{Case-Study: Binary String Parity }

 Many recent efforts have investigated relationships between Boolean functions (such as parity) and transformer architectures \cite{bhattamishra2023understanding,strobl2023transformers,bhattamishra2022simplicity,chiang2022overcoming} with a focus on which functions can be effectively represented  and learned by transformers.  This growing body of work provides a sound theoretical basis for this class of problems, the simplest of which we explore here for pedagogical and illustrative purposes.  These experiments can be easily replicated by the reader.

 A few facts about parity are:
 \begin{itemize}
     \item The parity of a finite binary string of any size is computed by adding the bits in the string and then noting whether the sum is even or odd.  If the sum is even, the parity is 0 and if the sum is odd, the parity is 1;
     \item The formal language defined by strings of even parity is a regular language accepted by a simple two state deterministic finite automaton.  As such it is one of the simplest nontrivial regular languages;
     \item Parity is extremely ``sensitive'' to perturbations in the input string because flipping any single bit in a string changes its parity;
     \item  It has been shown that transformer-based language models have difficulty learning the parity function for even moderately sized string lengths and generalization to longer length strings is poor
\cite{bhattamishra2023understanding,strobl2023transformers,bhattamishra2022simplicity,chiang2022overcoming};
\item Two reasonable properties to test for this task are accuracy (a simple metric) and monotonicity (a compound metric);
\item Technically speaking, both metrics are also second-order properties because they rely on an accurate computation of a string's parity but because this is done with a simple Python script that is easy to verify, we do not discuss this aspect further.
 \end{itemize}
 For these reasons, we have selected parity as an example to illustrate TEL'M reporting.  

  \subsubsection*{Training and Testing of Transformer-Based Parity Classifier}    
\begin{figure}[t] 
    \begin{center}
\includegraphics[width=6.5in]{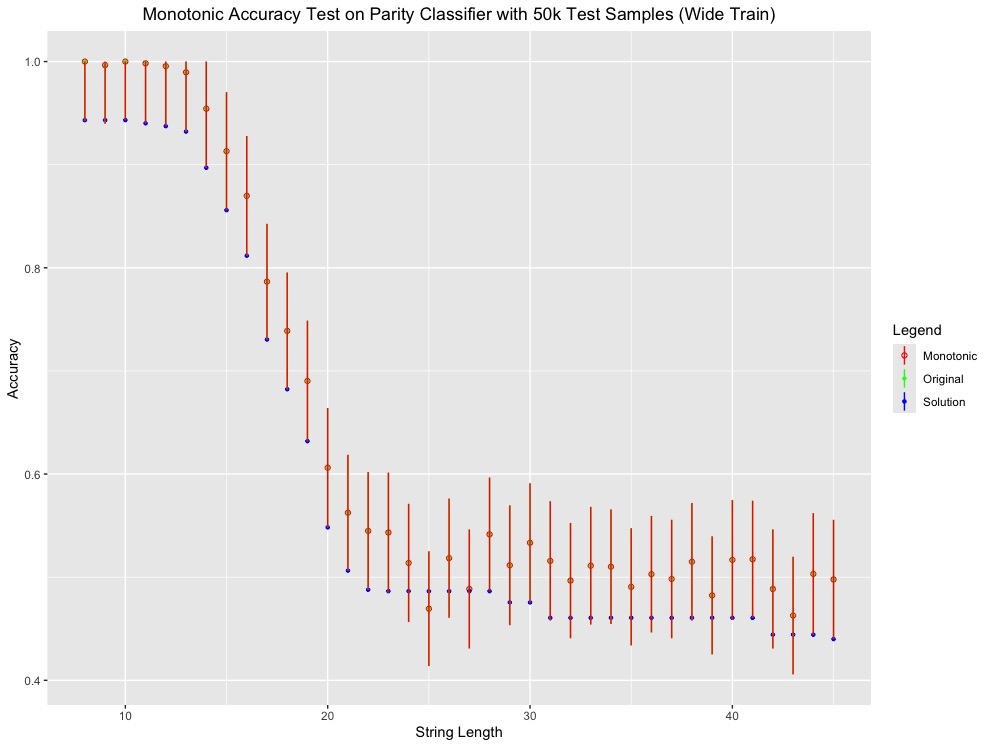}
\caption{ This plot shows the calculations behind the lower bound on distance to monotonic accuracy for the parity classifier trained on strings of length 8-45 ("wide train"). The total distance of this accuracy function from monotonic is lower bounded by $0$, as there is a solution within each confidence interval that is fully monotonic. The average value of $\epsilon$ in Hoeffding's inequality (equal to $\frac{1}{2}$ the width of the confidence interval for each string length, ignoring truncations at 0 and 1) is $0.057.$ As noted, the confidence that the true accuracy is in an interval is 99.9\%. } \label{WTS}
  \end{center}
\end{figure}

\begin{figure}[t] 
    \begin{center}
\includegraphics[width=6.5in]{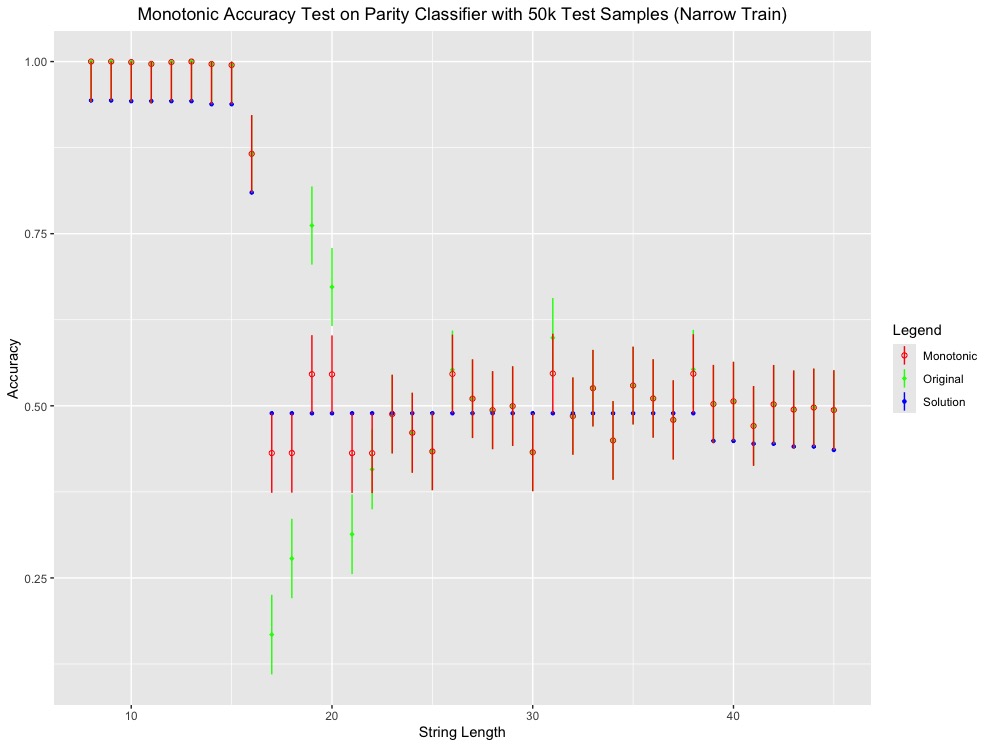}
\caption{ This plot shows the calculations behind the lower bound on distance to monotonic accuracy for the parity classifier trained on strings of length 8-15 ("narrow train"). The total distance of this accuracy function from monotonic is again lower bounded by $0.0254$. The average value of $\epsilon$ in Hoeffding's inequality is $0.057.$ As noted, the confidence that the true accuracy is in an interval is 99.9\%.} \label{NTS}
  \end{center}
\end{figure}

\begin{figure}[t] 
    \begin{center}
\includegraphics[width=6.5in]{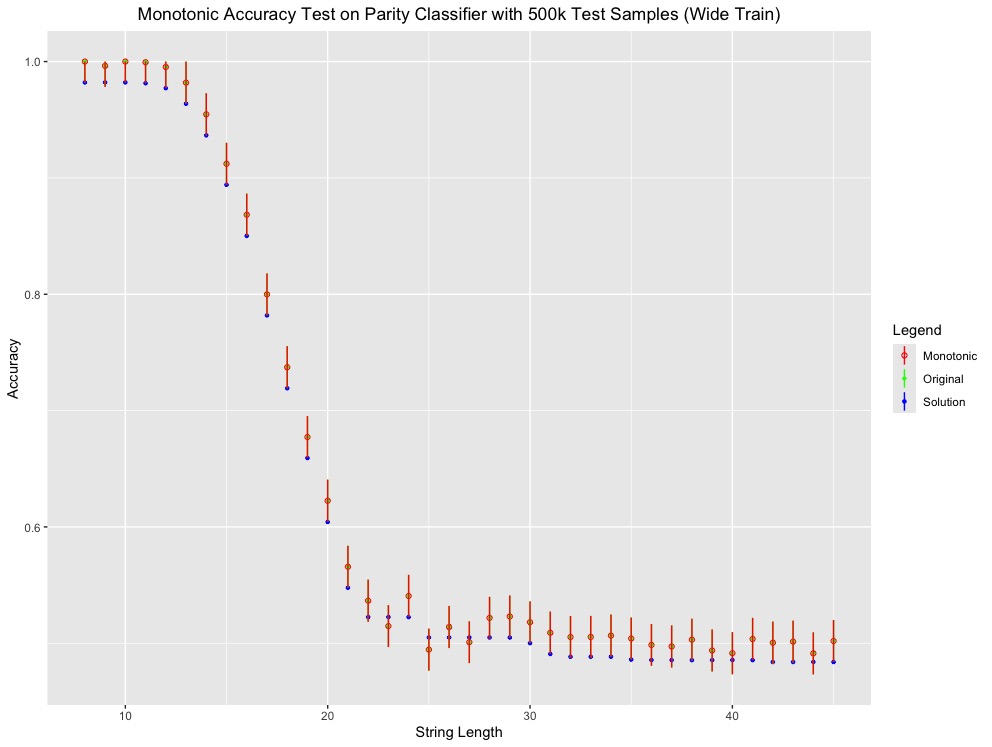}
\caption{ This plot shows the calculations behind the lower bound on distance to monotonic accuracy for the parity classifier trained on strings of length 8-45 ("wide train"). The total distance of this accuracy function from monotonic is lower bounded by $0$, as there is a solution within each confidence interval that is fully monotonic. The average value of $\epsilon$ in Hoeffding's inequality is $0.018.$ As noted, the confidence that the true accuracy is in an interval is 99.9\%. } \label{WTL}
  \end{center}
\end{figure}

\begin{figure}[t] 
    \begin{center}
\includegraphics[width=6.5in]{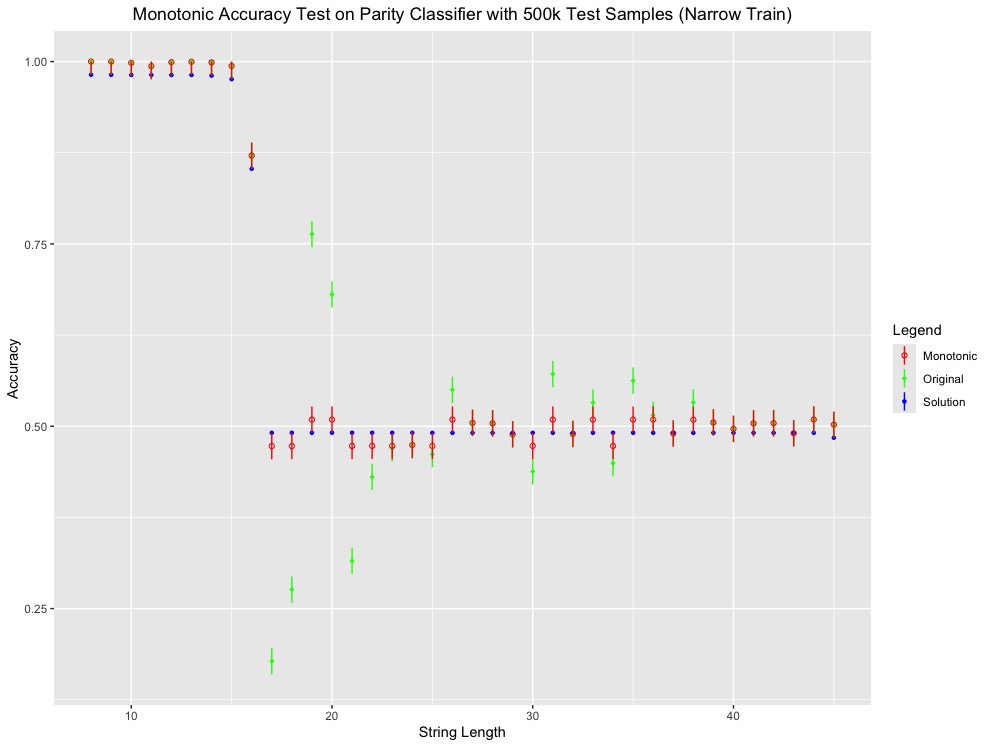}
\caption{ This plot shows the calculations behind the lower bound on distance to monotonic accuracy for the parity classifier trained on strings of length 8-15 ("narrow train"). The total distance of this accuracy function from monotonic is lower bounded by $0.0368$. The average value of $\epsilon$ in Hoeffding's inequality is $0.018.$ As expected, the more samples we take given the same desired confidence level, the narrower the confidence intervals, and the more we have to ``move around'' the confidence intervals in order to obtain a monotonic function. As noted, the confidence that the true accuracy is in an interval is 99.9\%.  } \label{NTL} 
  \end{center}
\end{figure}

   We trained and tested four parity classifiers based on the Chiang and Cholak Transformer  \cite{chiang2022overcoming}\footnote{https://github.com/ndnlp/parity}.  Our goal is to illustrate TEL'M suggested methodologies, terminology and reporting, not to study language model solutions to the parity problem.  The four models are:
   \begin{itemize}
       \item two trained on strings of length 8-15, which we call the ``Narrow Train'' classifier (test results are shown in Figures \ref{NTS} and \ref{NTL}), and;
       \item two trained on strings of length 8-45, which we call the ``Wide Train'' classifier (test results are shown in Figures \ref{WTS} and \ref{WTL}).
   \end{itemize}
   For each of these two groups, we performed testing using 1163 samples, as well as a test with about 10 times as many samples, so 11628 test samples for each length that we tested. We chose this approach to demonstrate the role of increased test sample size in getting smaller confidence intervals and tighter lower bounds on the distance to monotonicity. 
   
   Each of the four trained classifiers was tested on the parity task using strings of length between 3 and 45, although we limited our visualizations as well as our distance calculation to strings of length between 8 and 45. We did this for two reasons: first, since all classifiers performed nearly perfectly on very short strings, it does not affect the distance to monotonicity calculation. Second, these short strings are going to be very highly duplicated during testing, since we use over 1000 samples per string length which larger than the number of short length strings. For the smaller sample size test, we drew a total of 50,000 samples, averaging out to the 1163 samples per string length, while for the larger sample size test we drew 500,000 samples, or roughly 11628 samples per string length. During testing, the length of the string was selected uniformly between 8 and 45, and then a string was selected uniformly at random from all stirngs of that length. 

    The confidence interval derived from our tests as based on Hoeffding's Inequality, has a two-sided confidence level of $99.9\%$. Since there are $45-8+1 = 38$ confidence intervals, one for each string length tested, all of them hold simultaneously with probability $0.999^{38}  = 96.3\%$. 

    The four plots below show three confidence intervals for each of the 38 string lengths. The original sample averages and their confidence intervals are shown in green but many are occluded by red confidence intervals.  When a green confidence interval is not shown, it is beneath the red interval.
    
    Our linear programming approach to finding the minimal distance of these measurements to monotonicity are the shifted confidence intervals (shifted using one of the slack variables in the solution to the linear program) shown in red. Any string length that doesn't show green signifies that the original and shifted confidence intervals are completely overlapping, i.e. no shift was necessary (in the best-case) in order to establish monotonicity with its neighbors (and therefore overall). The distance between the green and red intervals is the``cost" incurred at that string length, or the amount that the entire interval would need to be shifted in order to contain a solution that is overall monotonic. Finally, the blue dots represent the actual linear program solutions, which is a point within the (possibly) shifted confidence interval that achieves monotonicity. 

    Note that some of the plots indicate a solution that decreases seemingly unnecessarily compared to its previous neighbor. The solution for the last four string lengths (42,43,44,45) in Figure \ref{WTS} could be shifted up slightly, and as long as they are below the average at string length 41, the entire function would still be monotonic. This is a reflection of the fact that the solution to the linear programming problem is not unique. This non-uniqueness does not affect the lower bound on the distance in any way, and the fact that the plots show the lowest possible sample averages that attain the minimum shows an arbitrary minimal solution, and an artifact of the lpSolve package in R.


 \newpage{\small
\begin{center}
\begin{table}[ht]
  \begin{tabular}{|l|p{3cm}|p{9cm}|} \hline 
    \textbf{TEL'M Template} & \multicolumn{2}{c|}{ Prepared by Paul Lintilhac (paul.s.lintilhac.th@dartmouth.edu) April 8 2024} \\
    \hline
    & Name & {\em Chiang and Cholak Transformer (2 layers, 2 heads) \cite{chiang2022overcoming}
     https://github.com/ndnlp/parity}\\
     \cline{2-3}
    Language Model & Version& {\em GitHub, downloaded March 2024 } \\
    \cline{2-3}
    ~& Training details& {\em 50,000 epochs, 100 steps each, batch size 1 (total of 5,000,000 samples with possible duplicates.) Trained on strings of length 8-15 (narrow train) and 8-45 (wide train)} \\
    \cline{2-3}
    ~&Benchmarks Used in Training Data& {\em None.}\\
    \cline{2-3}
      ~& Fine-tuning details& {\em None. Model trained from scratch.}  \\
    ~ & (if any) & ~  \\
        \cline{2-3}
      ~& Adaptations& {\em Using Classification Head on CLS (classify token) position.}  \\
    ~ & (if any) & ~  \\
    \hline
Task tested     & Description & {\em Compute parity of a variable length binary string as in  Chiang et. al.  \cite{chiang2022overcoming}}  \\
    \cline{2-3}
    & Dependencies & {\em Integer remainder from Python 3.9.6 library.} (label = len([a for a in w if a == 1]) \% 2 == 1 )  \\
    \hline
      ~& Description & {\em The model has the sole purpose of computing the parity of bit strings of variable length.}\\
      \cline{2-3}
   
Property tested & Number of Samples & {\em 50,000 (larger interval) and 500,000 (smaller interval)} \\
         \cline{2-3}
         & Distribution of Samples & {\em Uniform over length and uniform within strings of that length.} \\
         \cline{2-3}
    ~& Testing Algorithm & {\em Lower bound on distance to monotonicity using a linear program.} \\
     \hline

       ~& Description & {\em We measure the distance to monotonicity under the distribution that selects a string length uniformly and then a uniform random string of that length.}\\
       \cline{2-3}
Property Metric & Type Used & {\em Compound}   \\
    \cline{2-3}
    ~& Distance Distribution & {\em Uniform over length and uniform within strings of that length.} \\
    \hline 
    ~& Name \& Location & {\em Mac M3 Pro Max in Shelburne, VT} \\
       \cline{2-3}
Test Infrastructure & Description & {\em Commercial laptop}   \\
         \cline{2-3}
    ~& Time used for testing & {\em Testing done in under 10 minutes.} \\
    \cline{2-3}
    ~& Post processing & {\em Tests conducted in Python 3.9.6, Linear Program and plots implemented in R} \\

      \cline{2-3}
    ~& Benchmarks & {\em None} \\
     ~ & used (if any) & ~  \\
     \cline{2-3}
      ~& Stochasticity \& Temperature & {\em None.} \\
      \hline   
    ~& Open source model & {\em Github for Chiang and Cholak Transformer: https://github.com/ndnlp/parity} \\
     
    \cline{2-3}
Reproducability & Open source training data & {\em None.  Training data randomly generated. }   \\
      \cline{2-3}
    ~& Open source testing data& {\em None.  Testing data randomly generated.} \\
    \hline
  \end{tabular}
  \caption{This is a draft TEL'M report of the tests described here.  It is proposed as a template for reporting test and evaluation results for general language models and even Ai systems.}
\end{table}
\end{center}
 }
\end{document}